\documentclass{article}

\usepackage{microtype}
\usepackage{graphicx}
\usepackage{subfigure}
\usepackage{booktabs} % for professional tables
\usepackage{multirow}
\usepackage{makecell}
\usepackage{etoolbox}

\AtBeginEnvironment{tabular}{\scriptsize}

\usepackage{hyperref}

\usepackage[accepted]{icml2025}

\usepackage{amsmath}
\usepackage{amssymb}
\usepackage{mathtools}
\usepackage{amsthm}

\usepackage[capitalize,noabbrev]{cleveref}

\theoremstyle{plain}

\theoremstyle{definition}

\theoremstyle{remark}

\usepackage[textsize=tiny]{todonotes}

\icmltitlerunning{Rethinking AI/ML for Air Interface in Wireless Networks}

\begin{document}

\twocolumn[
\icmltitle{Position Paper: Rethinking AI/ML for Air Interface in Wireless Networks}

% It is OKAY to include author information, even for blind
% submissions: the style file will automatically remove it for you
% unless you've provided the [accepted] option to the icml2025
% package.

% List of affiliations: The first argument should be a (short)
% identifier you will use later to specify author affiliations
% Academic affiliations should list Department, University, City, Region, Country
% Industry affiliations should list Company, City, Region, Country

% You can specify symbols, otherwise they are numbered in order.
% Ideally, you should not use this facility. Affiliations will be numbered
% in order of appearance and this is the preferred way.
\icmlsetsymbol{equal}{*}

\begin{icmlauthorlist}
\icmlauthor{Georgios Kontes}{iis}
\icmlauthor{Diomidis S. Michalopoulos}{nokia}
% \icmlauthor{Mohammad Alawieh}{iis}
\icmlauthor{Birendra Ghimire}{iis}
\icmlauthor{Christopher Mutschler}{iis}
\end{icmlauthorlist}

\icmlaffiliation{iis}{Fraunhofer Institute for Integrated Circuits IIS, Fraunhofer IIS, N\"urnberg, Germany}
\icmlaffiliation{nokia}{NOKIA, Munich, Germany}

% \icmlaffiliation{prl}{Pattern Recognition Lab, Friedrich-Alexander-Universit\"at Erlangen-N\"urnberg, Erlangen, Germany}

\icmlcorrespondingauthor{Christopher Mutschler}{christopher.mutschler@iis.fraunhofer.de}

% You may provide any keywords that you
% find helpful for describing your paper; these are used to populate
% the "keywords" metadata in the PDF but will not be shown in the document
\icmlkeywords{Machine Learning, 3GPP standardization, Air Interface, 6G}

\vskip 0.3in
]

% this must go after the closing bracket ] following \twocolumn[ ...

% This command actually creates the footnote in the first column
% listing the affiliations and the copyright notice.
% The command takes one argument, which is text to display at the start of the footnote.
% The \icmlEqualContribution command is standard text for equal contribution.
% Remove it (just {}) if you do not need this facility.

\printAffiliationsAndNotice{}  % leave blank if no need to mention equal contribution
% \printAffiliationsAndNotice{\icmlEqualContribution} % otherwise use the standard text.

\begin{abstract}
AI/ML research has predominantly been driven by domains such as computer vision, natural language processing, and video analysis. In contrast, the application of AI/ML to wireless networks, particularly at the air interface, remains in its early stages. Although there are emerging efforts to explore this intersection, fully realizing the potential of AI/ML in wireless communications requires a deep interdisciplinary understanding of both fields.
%With the advent of 5G and the development of 6G, mobile communication systems offer fertile ground for impactful AI/ML-driven innovations, especially in the design and optimization of the Radio Access Network (RAN).
We provide an overview of AI/ML-related discussions in 3GPP standardization, highlighting key use cases, architectural considerations, and technical requirements. We outline open research challenges and opportunities where academic and industrial communities can contribute to shaping the future of AI-enabled wireless systems.
\end{abstract}

\section{Introduction}
\label{sec:intro}

In advanced 5G and future 6G systems, AI/ML will be essential to overcome challenges related to directional transmissions, rapidly varying channels, high feedback overhead and precise localization. For instance, AI/ML beam management~\cite{xue2024ai} leverages historical spatio-temporal data to swiftly predict optimal directed beams from a base station (BS) towards a user equipment (UE). This eliminates exhaustive beam search procedures and ensures that both base stations and UEs can quickly adjust to dynamic channel conditions. In addition, recurrent neural networks and transformer architectures address the problem of channel state information (CSI) prediction~\cite{jiang2025ai} (i.e., the inherent delays between the estimation of channel state and its usage) by forecasting future channel states, ensuring timely and accurate link adaptation even in high-mobility scenarios. Complementing these methods, deep learning-based encoder/decoder architectures learn concise latent-space representations of channel data to compress the CSI~\cite{lin2025bridge}, which is periodically reported from the UE to the BS. This leads to lower latency and improved spectral and energy efficiency. Finally, AI/ML positioning~\cite{alawieh20235g}, enhances the reliability of location-based services, by extracting relevant-only channel features and adapting across different deployment scenarios. 

Collectively, this holistic AI/ML framework has been debated and shaped within 3GPP during Releases 18 and 19~\cite{tr38843}. Although the 3GPP standardization task force has laid the foundations for the adoption of AI/ML in the air interface, the development and deployment of both performing and cost-efficient AI/ML models come with their own challenges. As we enter the 6G era in standardization, where AI/ML is meant to play a central role, these challenges must be discussed from day one.

%Section~\ref{sec:open_issues} discusses the additional complexity that AI / ML adoption brings to the standardization of the air interface of wireless radio networks, Section~\ref{sec:way_forward} presents a number of forward-looking approaches to address these challenges from day one in the new 6G standard, while~\ref{sec:conclusion} concludes this position paper.

\section{Current Discussion \& Open Challenges}
\label{sec:open_issues}

Any AI/ML application at scale comes at a high maintenance cost~\cite{ashmore2021assuring}. The availability of a clean and automated pipeline that facilitates all three phases of data governance (Sec.~\ref{sec:data}), model training and testing, and model deployment (including model monitoring and management) (Sec.~\ref{sec:LCM}) in both a central and distributed manner (Sec.~\ref{sec:interoperability}) has proven to be equally (or more) important as the model's capabilities and performance.

For wireless networks, things are even more complicated. Here, the definition and implementation of such a pipeline for AI/ML integration are also hindered by the number of different stakeholders that need to come to an agreement. The air interface facilitates a complex interplay between UE, BS and core network (NW) vendors (constrained also by the needs and requirements of the mobile network operators) that need to agree on common and possibly synergetic solutions for all stages of the AI/ML model pipeline, while avoiding any proprietary information being exposed.

% This section highlights the main challenges a successful AI/ML integration in the wireless air interface must face.

\subsection{Data Governance}
\label{sec:data}

Data collection, management (data cleaning, privacy, standardization, enhancement, availability, etc.), and security (rejection of malicious/manipulated data) for model training, testing, and monitoring are core requirements for AI/ML-based solutions~\cite{huang2024data}. The labeled data required for model training is usually owned by the organization that trains the model. It is similar for monitoring: once the model is deployed, usually only the model owner has access to the relevant key performance indicators (KPIs).

\textbf{Challenges.} In current 3GPP discussions, the fundamental issue of which data can be collected in a way that respects user's security and privacy aspects plays central role. In parallel, the key questions of data ownership and access mechanisms still remain unanswered. For instance, in positioning, a UE-sided model requires area-specific labels (its true position, provided by the network) to be used for training, fine-tuning or monitoring. The situation is similar for NW-sided models. If we consider NW-sided beam management, the BS has full control on the transmitted beams but requires UE-sided beam measurements for a proper (training, testing, or monitoring) dataset.

\subsection{Life Cycle Management (LCM)}
\label{sec:LCM}

\subsubsection{Model Complexity vs Generalization}
\label{sec:complexity}

The energy needed for the training and inference operations of high-performing models increases with model size, as larger models exhibit better performance and generalization capabilities~\cite{brutzkus2019why}. Although UE and NW energy savings strategies are central within 3GPP discussions, the implications of using AI/ML models on the energy footprint remain largely underexplored. In fact, fulfilling several requirements set forth in Release 19 AI/ML for air interface would result in larger and more complex model architectures and thus increased energy consumption for both the UEs and the NW. Examples include: (i) training with larger/mixed datasets, which results in larger AI/ML models; (ii) switching between smaller (even site-specific) models, which necessitates additional signaling between the UE and the NW, and switching decision mechanism, and (iii) fine-tuning, which depends on a pre-trained (and thus potentially large) model.

\textbf{Challenges.} Robust generalization of UE-sided models requires tackling diverse radio environments (e.g., channel conditions and interference) that quickly degrading performance if models are not adapted. Training or fine-tuning one large or multiple specialized models places heavy energy and resource demands on UEs and requires detection of distribution shifts for unseen or rare scenarios. For smaller, specialized models, model switching or individual model update introduces additional complexity and latency as UEs and NW nodes must streamline when to switch or retrain.

\begin{table*}[t]
\caption{Promising research directions to unlock the adoption of AI/ML in air interface within 6G.}
\label{tb:sol}
%\vskip 0.15in
\begin{center}
\begin{small}
\begin{sc}
\begin{tabular}{lccccccc}
\toprule
\rule{0pt}{2ex} \multirow{4}{*}{\makecell[l]{Research\\ Direction}} & \multicolumn{7}{c}{Challenges}\\
\cline{2-8}
 \rule{0pt}{4ex} & \multirow{3}{*}{\makecell[l]{Data\\ Governance}} & \multicolumn{5}{c}{Model Life Cycle Management} & \multirow{3}{*}{\makecell[l]{Interoper-\\ability}}\\
\cline{3-7}
  \rule{0pt}{4ex} & & \makecell[l]{Monitoring \&\\  Management} & \makecell[l]{Model\\ Complexity} & \makecell[l]{Genera-\\lization} & \makecell[l]{KPI\\ alignment} & \makecell[l]{Simplified\\ Testing} & \\
\midrule
\rule{0pt}{4ex} \makecell[l]{Multi-task learning}  & \checkmark  & \checkmark  &   & \checkmark  &   &   & \checkmark  \\
\rule{0pt}{4ex} \makecell[l]{Conditional architectures}  &   & \checkmark  & \checkmark  &   &   &   & \checkmark  \\
\rule{0pt}{4ex} \makecell[l]{Root Cause analysis}  & \checkmark  & \checkmark  &   &   &   & \checkmark  & \checkmark  \\
\rule{0pt}{4ex} \makecell[l]{Opportunistic data collection}  & \checkmark  & \checkmark  & \checkmark  &  &   & \checkmark  &   \\
\rule{0pt}{4ex} \makecell[l]{Self-supervised/ Meta/Active\\ learning}  & \checkmark  & \checkmark  &  & \checkmark  &   &   &   \\
% \rule{0pt}{4ex} \makecell[l]{Meta\\ learning}  & \checkmark  &   &   & \checkmark  &   &   &   \\
% \rule{0pt}{4ex} \makecell[l]{Active\\ learning}  & \checkmark  & \checkmark  &   & \checkmark  &   &   &   \\
\rule{0pt}{5ex} \makecell[l]{Reinforcement Learning \& \\ AI-based  Optimization}  & \checkmark  &   & \checkmark  & \checkmark  & \checkmark  &   & \checkmark  \\
\bottomrule
\end{tabular}
\end{sc}
\end{small}
\end{center}
\vskip -0.1in
\end{table*}

\subsubsection{Model Testing}
\label{sec:testing}

One of the most important steps prior to model deployment is model testing. Different types of test can be implemented, ranging from the evaluation of prediction accuracy on a holdout data set to more elaborate testing methods that involve different versions of the model being evaluated by real users. Again, the organization that owns the model usually has all test data available and designs test protocols.

\textbf{Challenges.} Defining testing procedures for all use cases is not straightforward (e.g., there are no agreed testing scenarios for positioning methods that track moving UEs). There are two main constraints: (i) designing tests that are universal for the AI/ML models developed by all vendors at every UE type, and (ii) defining tests for fine-tuned or area-specific models. The former constraint requires widely applicable and simpler testing mechanisms, while the latter opens the discussion on on-device post-deployment testing protocols.

\subsubsection{Model monitoring and management}
\label{sec:testing}

%Model monitoring and management is crucial for real-world applications.
Monitoring analytics are diverse: From simple (hardware) system performance metrics (e.g., memory utilization) and model input/output monitoring for detecting out-of-distribution inputs/outputs, to time-consuming and cost-inducing label-based monitoring approaches. Usually, the detection of model performance drop triggers a new data collection and model retraining (or fine-tuning) procedure.

\textbf{Challenges.} As also mentioned above, monitoring label availability (for label-based monitoring analytics) requires alignment between UE and NW sides, which blurs the ownership status of the monitoring dataset. Second, even though for UE-(BS-)sided models model monitoring and management are handled by the UE (BS), the NW is still responsible for the performance in an area. This implies that the NW can request the UE (or BS) to start a monitoring session on their respective models to determine if performance is adequate or a fallback to a different approach is required. What further complicates things is that reporting exact monitoring KPIs might  result in security issues: If a malicious/adversarial UE reports wrong KPIs to the NW, the NW cannot easily detect this, while if the UE reports its model predictions to the NW, there is a danger of model extraction~\cite{tramer2016stealing, oliynyk2023know}.

\subsubsection{Performance KPIs}
\label{sec:kpis}

Performance KPIs for AI/ML model training, testing, and monitoring consider the perfect predictive performance of the model in all use cases: beam management strives to identify the beam with the highest RSRP among top predicted beams, or predicting the correct RSRP for all beams; CSI prediction is tasked to forecast future channel conditions perfectly; CSI compression aims at reconstructing CSI flawlessly at the BS; and positioning is expected to estimate UE positions at millimeter accuracy.

\textbf{Challenges.} Such stringent prediction-accuracy-oriented KPIs are not well aligned (see~\citet{russell2020artificial} for a definition of alignment) to the real world applications. For example, video streaming requires different beam quality than text communication, many positioning tasks do not require millimeter or centimeter accuracy, and for some tasks incomplete CSI information is enough~\cite{jiang2025ai}. The challenge here, in addition to identifying the right KPI for each task, is to train and deploy models that strike the right balance between complexity and performance requirements, as posed by the underlying application.

\subsection{Model Interoperability}
\label{sec:interoperability}

Many scenarios require ``two-sided'' models that are deployed in a distributed fashion between UE and BS, which involve a joint operation of the models at both ends. In addition to low complexity, AI/ML models should strive for a robust exchange of information between the entities, enabling a coordinated operation of the model as a whole. %%targeting an optimized utilization of models that are distributed between devices and base stations.

\textbf{Challenges.} The AI/ML model at UE- and NW-side may come from different providers, facing the challenge of interoperability. Vendors at each side need to support the model yet without disclosing their respective proprietary models. This implies that UE and NW vendors must architect sophisticated interoperability frameworks that facilitate selective sharing of their models. This would ensure the preservation of complete model confidentiality while fostering the harmonious and successful operation of two-sided models.

\section{%Promising 
Future Research Directions}
\label{sec:way_forward}

We discuss promising research directions that support true integration of AI/ML in wireless air interfaces (Tab.~\ref{tb:sol}).
%By addressing key challenges—from data governance to energy constraints—this discussion introduces innovative approaches, including multi-task learning, conditional network architectures, and reinforcement learning strategies. These directions not only promise enhanced system-level performance but also pave the way for adaptable, robust, and energy-efficient AI/ML solutions in the upcoming 6G era.

\subsection{Multi-Task Learning}

AI/ML use cases are typically developed isolated to one another, which results in high development cost ~\cite{jiang2025ai}. Possibly overlapping data is collected separately and thus trained models encode redundant information. To address this, multi-task learning~\cite{zhang2021survey} deploys a single backbone model as a common representation of the channel (e.g., see~\citet{ott2024radio}), which in sequence provides targeted ``heads'' tailored to the involved use cases. 

%By consolidating tasks, multi-task learning inherently reduces the overall demand for data requirements for training, testing, and monitoring, since instead of having separate, duplicate data pipelines for each task, shared data streams can ensure more systematic collection.

This also facilitates model monitoring and management, as maintenance of a single multi-task model is more straightforward than tracking multiple specialized models, while retraining or fine-tuning can be done within a unified pipeline. As task representations share latent features, fine-tuning (or even adding a new use case / model head) can require fewer data, since such shared knowledge can be effectively utilized. Furthermore, if complementary tasks on the UE side (e.g., beam management and resource scheduling, respectively) partially share the backbone model, the integration of two-sided training or inference routines in multi-task models is facilitated as well, thus enabling model interoperability.

\subsection{Conditional Neural Architectures}

Even with multi-task approaches, universal ``one-size-fits-all'' models may still be suboptimal for a number of UEs. Some UEs frequently encounter a narrow set of channel conditions (e.g., repeated commuting routes), therefore, on-device personalized sub-models can yield substantial gains, like increased performance and reduced signaling overhead.  

A way to tackle this could be the adoption of two-sided models, enhanced with early exit architectures~\cite{teerapittayanon2016branchynet}. This would allow earlier termination when prediction confidence is high, exploiting the fact that many data samples can be classified using only the earlier layers of the model. For challenging conditions (e.g., a shift from LOS to multipath NLOS), the inference can continue through additional layers, ensuring uninterrupted performance. 

Some of the neural network layers can be shared between the UE and NW vendors, allowing flexible execution that eases the UE-side computational burden and ensures a high level of interoperability between the distributed sub-models, while protecting the proprietary properties of the UE- and NW-side parts of the model. This way, (distributed) model monitoring and management is also more principled, as performance degradation can be isolated to specific exits, instead of re-verifying the entire network.

\subsection{Root Cause Analysis}

When an AI / ML model performs poorly, the monitoring entity should not only report a failure but also pinpoint the source of the problem. For instance, as studied within 3GPP Release 19, if the performance of a two-sided (encoder/decoder) CSI compression model degrades, an investigation on whether the encoder, the decoder, or a change in the radio environment is responsible enhances the interoperability properties of such distributed models. 

Such information also allows making informative decisions on how to mitigate the effect of a detected issue, e.g., retrain only part of the model, switch to a smaller fallback model, or temporarily deactivate the AI/ML functionality. This enables more transparent model testing procedures and can even support high-level root-cause explanations, e.g., using natural language~\cite{roy2024exploring, manjunath2025multimodal}.

\subsection{Opportunistic Data Collection}

Instead of continuously collecting positioning labels or logging CSI and beam measurements, UEs and base stations can report data only when certain triggers appear. For instance, the UE or the NW might request monitoring data only if the UE’s performance abruptly changes -- a sign that the current environment differs from what the model has seen before. Likewise, for obtaining the true labels when legacy positioning methods are unreliable, the NW can opportunistically activate external sensors or measurements to acquire secondary information to obtain labels in challenging scenarios (e.g., using camera or other UE sensors). 
% This selective data-reporting approach allows effective fine-tuning or monitoring without massive overhead, since the UE collects and transmits extra data only when encountering such landmarks. 

Such targeted, trigger-based data logging (for training, testing and monitoring) not only enables a more agile model monitoring and management framework, but also unlocks the ability for on-demand testing in highly variable scenarios, thus allowing effective use of test and label resources instead of continuous, exhaustive logging.

\subsection{Reinforcement Learning / Optimization-Based AI}
 
AI/ML studies within 3GPP standardization prioritize improving a single metric (usually prediction accuracy), often without considering the broader objectives tied to system-wide KPIs, i.e., without involving Quality of Service (QoS) and Quality of Experience (QoE) requirements. For example, simply increasing channel prediction accuracy does not necessarily translate into the best end-to-end performance for tasks such as positioning, beam management, or channel state feedback~\cite{jiang2025ai}.

To address this gap, future AI/ML designs should include a family of online learning or optimization-based approaches, such as reinforcement learning~\cite{sutton2018reinforcement}, recommender systems~\cite{zhang2019deep} or mixed-integer optimization~\cite{wolsey1999integer}. These adaptive algorithms, unlike fixed models solely optimized for higher prediction accuracy, can be combined with task-oriented KPIs and are able to continuously adjust to real-world conditions, such as network load fluctuations, device mobility, or localized interference patterns. 

Unsurprisingly, such methods also ease the burden on data collection for model (called policy within RL) training, testing, and monitoring, since stringent requirements on high-quality labeled datasets no longer exist. Instead, task-related measurements used to calculate reward (and possibly a set of constraints) estimates must be collected. This also enhances UE- and NW-side interoperability, since each side can maintain its own policy function and still coordinate via reward exchanges or partial state observations.

\subsection{Efficient use of labeling resources}

Most of the widely-adopted AI/ML approaches and model architectures can benefit from algorithms that enable efficient use of labeling resources, such as self-supervised~\cite{liu2021self}, meta-learning~\cite{hospedales2021meta} and active learning~\cite{ren2021survey} techniques. For example, in several scenarios, a meta-learned model can quickly adapt when deployed to new cell sites or applied under changed channel conditions. In CSI prediction/compression use cases, self-supervised methods can process channel measurements to predict future channel behavior or compress CSI data, without any dependency on manual annotation. Or in positioning, methods such as channel charting~\cite{studer2018channel} can alleviate the dependency on positioning reference units for the availability of accurate position labels.

Complementary, active learning enables on-demand model monitoring, testing, and fine-tuning, ensuring that only the most informative data samples (e.g., when model prediction uncertainty is high or it starts to exhibit degraded performance) are collected and are used for model adjustments.

\section{Conclusion}
\label{sec:conclusion}

There is no single silver bullet to address the aforementioned challenges. Future research should focus on hybrid solutions that combine modular, adaptive algorithms with robust data governance and continuous model monitoring, aiming to reduce the reliance on hard-to-acquire and cost-deficient label-based approaches.

\clearpage

\bibliography{references}
\bibliographystyle{icml2025}

%%%%%%%%%%%%%%%%%%%%%%%%%%%%%%%%%%%%%%%%%%%%%%%%%%%%%%%%%%%%%%%%%%%%%%%%%%%%%%%
%%%%%%%%%%%%%%%%%%%%%%%%%%%%%%%%%%%%%%%%%%%%%%%%%%%%%%%%%%%%%%%%%%%%%%%%%%%%%%%
% APPENDIX
%%%%%%%%%%%%%%%%%%%%%%%%%%%%%%%%%%%%%%%%%%%%%%%%%%%%%%%%%%%%%%%%%%%%%%%%%%%%%%%
%%%%%%%%%%%%%%%%%%%%%%%%%%%%%%%%%%%%%%%%%%%%%%%%%%%%%%%%%%%%%%%%%%%%%%%%%%%%%%%
% \newpage
% \appendix
% \onecolumn
% \section{You \emph{can} have an appendix here.}

% You can have as much text here as you want. The main body must be at most $8$ pages long.
% For the final version, one more page can be added.
% If you want, you can use an appendix like this one.  

% The $\mathtt{\backslash onecolumn}$ command above can be kept in place if you prefer a one-column appendix, or can be removed if you prefer a two-column appendix.  Apart from this possible change, the style (font size, spacing, margins, page numbering, etc.) should be kept the same as the main body.
%%%%%%%%%%%%%%%%%%%%%%%%%%%%%%%%%%%%%%%%%%%%%%%%%%%%%%%%%%%%%%%%%%%%%%%%%%%%%%%
%%%%%%%%%%%%%%%%%%%%%%%%%%%%%%%%%%%%%%%%%%%%%%%%%%%%%%%%%%%%%%%%%%%%%%%%%%%%%%%

\end{document}